\documentclass[11pt,a4paper]{article}
\usepackage{times,latexsym}
\usepackage{url}
\usepackage[T1]{fontenc}

\usepackage[acceptedWithA]{tacl2018v2}
\usepackage[]{tacl2018v2}

\usepackage{amsmath,amsthm,amsfonts,amssymb,bm}
\usepackage{xspace,mfirstuc,tabulary}
\usepackage{booktabs}
\usepackage{tabularx}
\usepackage{microtype}      
\usepackage{comment}        
\usepackage{amsopn}
\usepackage{xcolor}
\usepackage{multirow}
\usepackage{tikz,pgfplots}
\usepackage{array}
\usepackage{colortbl}
\usepackage{hhline}
\usepackage{makecell}
\usepackage{enumitem}
\usepackage{soul}
\usepackage{tikz,pgfplots}
\usepgfplotslibrary{groupplots}
\pgfplotsset{compat=1.14}
\usetikzlibrary{shapes,arrows, positioning}

\definecolor{g-red}{HTML}{DB4437}
\definecolor{g-blue}{HTML}{4285F4}
\definecolor{g-green}{HTML}{0F9D58}
\definecolor{g-yellow}{HTML}{F4B400}
\definecolor{g-orange}{HTML}{FF9800}
\definecolor{g-grey}{HTML}{9E9E9E}

\newif\iftaclinstructions
\taclinstructionsfalse 
\iftaclinstructions
\renewcommand{\confidential}{}
\renewcommand{\anonsubtext}{(No author info supplied here, for consistency with
TACL-submission anonymization requirements)}
\newcommand{\instr}
\fi

\iftaclpubformat 

\else

\fi

\newcommand{\defeq}{\overset{{\scriptscriptstyle\text{def}}}{=}}
\newcommand{\slices}{\mathtt{slice_s}}
\newcommand{\pextrap}{p_{\scriptscriptstyle{\mathtt{Ex2}}}}

\newcommand{\totext}{\mathtt{to\_text}}
\newcommand{\fromtext}{\mathtt{from\_text}}

\newcommand{\simwr}{\overset{{\scriptscriptstyle \text{wr}}}{\sim}}
\colorlet{light cyan}{cyan!33}
\colorlet{light green}{green!33}
\colorlet{dark green}{green!33!black}
\newcommand{\cspan}[3][yellow]{\begingroup\sethlcolor{#1}[\hl{#2} #3 ]\endgroup}
\newcommand{\mhl}[2][yellow]{\mathchoice%
  {\colorbox{#1}{$\displaystyle#2$}}%
  {\colorbox{#1}{$\textstyle#2$}}%
  {\colorbox{#1}{$\scriptstyle#2$}}%
  {\colorbox{#1}{$\scriptscriptstyle#2$}}}%
\newcommand{\sep}{$~|~$} 
\newcommand{\nl}[1]{{\fontfamily{cmtt}\selectfont``#1''}} 

\title{Neural Data Augmentation via Example Extrapolation}

\author{
Kenton Lee $^*$\hspace{0.15em} Kelvin Guu $^*$\hspace{0.15em} Luheng He $^*$\hspace{0.15em} Timothy Dozat $^*$\hspace{0.15em} Hyung Won Chung
 \Thanks{Equal contribution from all authors.} \vspace{0.5em}\\
 Google Research \vspace{0.5em}\\
 {\texttt{\{kentonl, kguu, luheng, tdozat, hwchung\}@google.com}} \\
}

\date{}

\begin{document}
\maketitle
\begin{abstract}
In many applications of machine learning, certain categories of examples may be underrepresented in the training data, causing systems to underperform on such ``few-shot'' cases at test time. A common remedy is to perform data augmentation, such as by duplicating underrepresented examples, or heuristically synthesizing new examples. But these remedies often fail to cover the full diversity and complexity of real examples.

We propose a data augmentation approach that performs neural \textbf{Ex}ample \textbf{Ex}trapolation (Ex2). Given a handful of exemplars sampled from some distribution, Ex2 synthesizes new examples that also belong to the same distribution.
The Ex2 model is learned by simulating the example generation procedure on data-rich slices of the data, and it is applied to underrepresented, few-shot slices.

We apply Ex2 to a range of language understanding tasks and significantly improve over state-of-the-art methods on multiple few-shot learning benchmarks, including for relation extraction (FewRel) and intent classification + slot filling (SNIPS).
\end{abstract}


\section{Introduction}\label{sec:intro}

\begin{figure}[hbt!]
\includegraphics[width=\columnwidth]{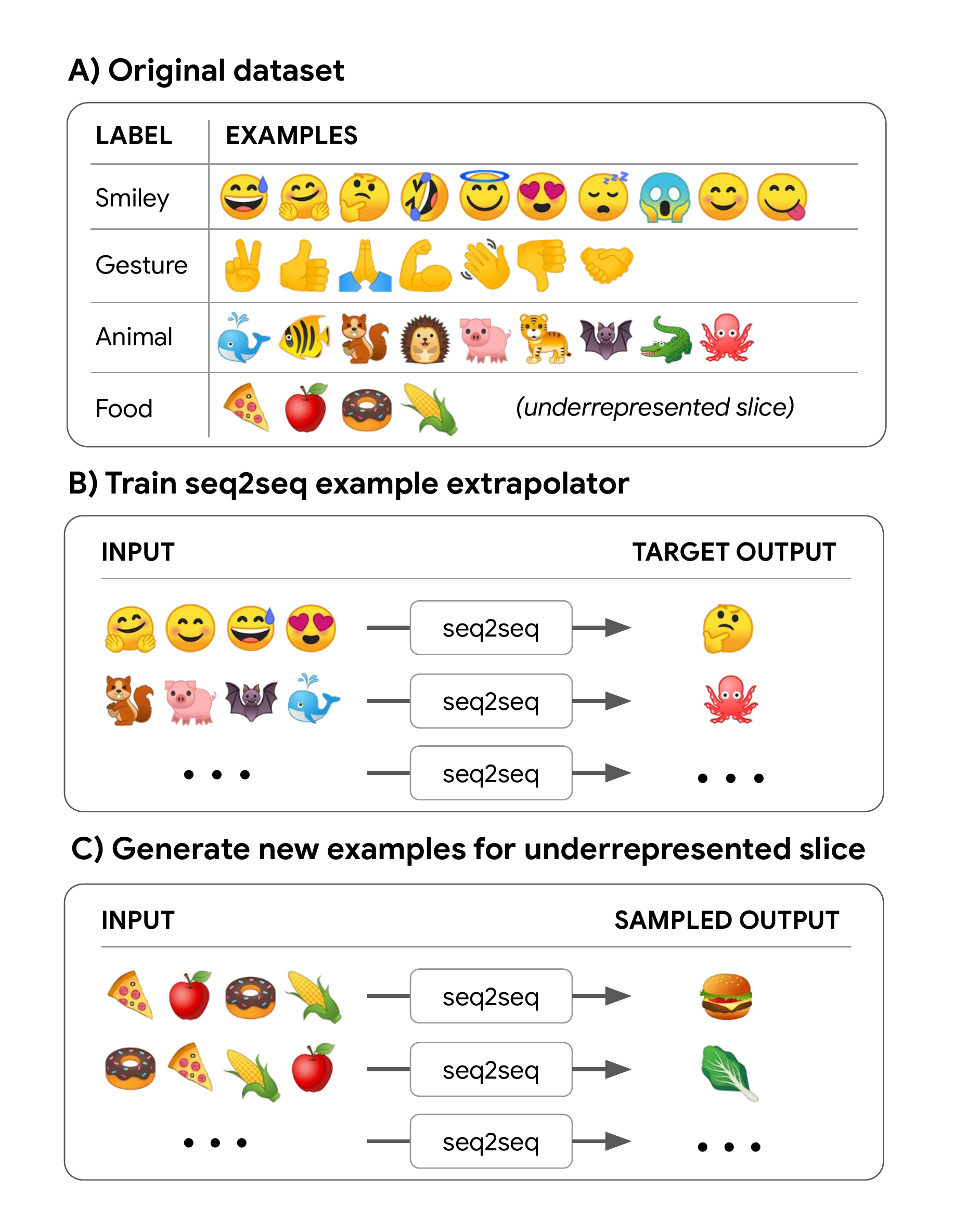}
\label{fig:intro}
\vspace{-2.5em}
\caption{Illustration of our approach (each emoji represents a data point). We first group our training data into different slices and identify slices that are underrepresented (A). Then we train an example extrapolator, which takes several examples from the same slice as input, and learns to synthesize a new example belonging to the same slice (B). Finally, we use the extrapolator to synthesize new examples for underrepresented slices of the dataset (C).}
\end{figure}

Data collection is a noisy process, and there are often significant mismatches between training and test distributions, leading to certain slices of data being underrepresented in the training set.
For example, developers of a dialog agent may regularly add new ``intents'' to their system's set of capabilities, but data collection for each new intent often lags behind~\cite{bapna2017towards,gaddy2020overcoming}. More generally, this issue can be a chronic problem for tasks with constantly expanding output spaces, such as relation extraction~\cite{han2018fewrel} and entity linking~\cite{logeswaran2019zero}, or particularly long-tail output spaces, such as fine-grained image classification~\cite{akata2015evaluation}. In such situations, existing systems can severely underperform on underrepresented slices of the data due to the incorrect prior probability of predicting them.

Data augmentation is a popular solution for biased or imbalanced data, either by duplicating examples or using heuristics to synthesize new examples~\cite{perez2017effectiveness}. However these heuristics may not scale well and are poor approximations of the complexity of real examples.

In this paper, we propose an approach for learned data augmentation that uses a neural \textbf{Ex}ample \textbf{Ex}trapolator (Ex2) to synthesize new examples (illustrated in Figure \ref{fig:intro}). Ex2 takes as input a handful of examples (``exemplars'') drawn from an underrepresented slice of data and learns to synthesize new examples that fall within the same slice and distribution as the exemplars (Step C in Figure~\ref{fig:intro}).
Ex2 learns to extrapolate to new examples by simulating this procedure using random subsets of the training data that already have a large number of examples (Step B in Figure~\ref{fig:intro}).

Our approach has strong connections to several recent works on using language models for data augmentation~\cite{kumar2019closer} and zero-shot learning~\cite{Brown2020LanguageMA}, as well as methods for few-shot learning via nearest-neighbor models~\cite{snell-etal-2017-prototypical,vinyals-etal-2016-matching}. We discuss these connections at length in Section~\ref{sec:related}.

We apply Ex2 to several language understanding tasks that contain few-shot slices of data, including relation extraction (FewRel) and intent classification/slot-filling tasks (CLINC150 and SNIPS). By correcting for the underrepresentation of those slices with Ex2 data augmentation, we significantly improve state-of-the-art methods.
\begin{table*}[t!]
\setlength{\tabcolsep}{0.4em}
\centering
\scriptsize
\def\tabularxcolumn#1{m{#1}}
\begin{tabularx}{\linewidth}{lXXc}
\toprule
Task & Input sequence & Output sequence & \makecell{Anonymized\\slice identity} \\
\midrule
\makecell[l]{Relation \\ extraction}
& 
\nl{Their \cspan[light cyan]{1}{arrival} led to \cspan{0}{utter chaos}\sep
Light blue shows the extent of the \cspan{0}{flood} from \cspan[light cyan]{1}{rivers}}
& 
\nl{An \cspan{0}{oil spill} caused by \cspan[light cyan]{1}{a collision} closed the ship channel.} & 
\makecell{
$\begin{aligned}
\mathit{relation} &= \mathit{effect}\\
\mhl{\mathit{head}} &= 0\\
\mhl[light cyan]{\mathit{tail}} &= 1
\end{aligned}
$}\\
\cmidrule{1-4}
Classification 
&   \nl{Check my car's tire pressure 
    \sep Should I pump my tires 
    \sep What's the air level in my tires}  
&   \nl{Are my tires under-inflated} &
\makecell[l]{
$\mathit{intent}$\\
$~~= \mathit{tire\_pressure}$
}\\
\cmidrule{1-4}
Slot filling 
&   \nl{Weather in \cspan{0}{New Beaver} \sep
    What's the forecast for \cspan[light cyan]{1}{Dec  1st} \cspan{0}{in Keeneland}} &
    \nl{How will the weather be at \cspan{0}{Steven's Pass} \cspan[light cyan]{1}{this weekend}}& 
    \makecell{
$\begin{aligned}
\mathit{intent} &= \mathit{weather}\\
\mhl{\mathit{location}} &= 0 \\
 \mhl[light cyan]{\mathit{time}} &= 1
\end{aligned}
$}\\
\bottomrule
\end{tabularx}
\caption{Training examples for the Ex2 model. The examples are adapted/shortened from the training sets described in Section \ref{sec:experiments}. The anonymization and slicing strategies will also be described in Section \ref{sec:experiments}.} 
\label{tab:examples}
\end{table*}

\section{Approach}\label{sec:approach}

\subsection{Overview}
Throughout this paper, we focus on applying Ex2 to standard supervised learning tasks. Our approach consists of the following high-level steps:
\begin{enumerate}[noitemsep]
    \item Organize training data into multiple slices.
    \item Train an example extrapolator using data from those slices.
    \item Use the example extrapolator to generate new synthetic data for underrepresented slices of the dataset.
    \item Train a model on the union of the synthetic data and real data.
\end{enumerate}

The core of the approach is the example extrapolator, which is a generative model that aims to recover the full distribution of examples given only a few samples from that distribution.
During inference (Step C in Figure~\ref{fig:intro}), the extrapolator takes as input the concatenation of $K$ gold examples that come from an underrepresented slice of the dataset and generates new examples that belong to the same slice. To train this extrapolator (Step B in Figure~\ref{fig:intro}), we simulate this procedure by randomly selecting $K + 1$ gold examples from a data-rich slice and optimize the log-likelihood of one of the examples given the other $K$ examples.

The synthetic data sampled by performing inference on the underrepresented slices can then be combined with existing data, which is applicable to any supervised learning setup.

The rest of this section motivates and formalizes this approach.

\subsection{Formal Definitions}
We denote a training example as $e = (x, y)$, where $x$ is the input and $y$ the output. In a text classification task, for example, $x$ would be a snippet of text (e.g., ``Play a song''), and $y$ the class (e.g., \texttt{PlayMusic}).
\paragraph{Slicing data}
\label{sec:slicing}
In many tasks, there is a natural way to slice data into different subsets of interest. For example, a slice could be the set of all examples sharing a given label, or all examples in a particular language, or with a particular syntactic construction. Ex2 makes no assumptions about how data is sliced --- for any given application, it is up to the practitioner to slice the data in a way that exposes important but underrepresented slices, which Ex2 can then target for data augmentation. 

To formalize the notion of slicing, we assume that the practitioner defines a list of $S$ slicing functions, $\slices$ for $s = 1, \ldots, S$, where each function $\slices(e)$ is a Boolean function indicating whether example $e$ belongs in slice $s$ (potentially overlapping with other slice functions). For example, a text classification slicing function that groups all examples with the same label $c$ would be $\mathtt{slice}((x,y)) \defeq \delta(y = c)$.

Given a dataset $D$, we define the $s^{th}$ slice of that dataset to be $D_s \defeq \{e \in D \mid \slices(e) = \mathtt{true}\}$. For a set of slices $S$, we also define $D_S \defeq \displaystyle \bigcup_{s \in S} D_s$.

\paragraph{Few-shot versus many-shot.}
We will assume that underrepresented slices have only a few examples each, so we refer to these as \textit{few-shot slices} (denoted as $F$): we will perform data augmentation for these slices. We call the remaining slices \textit{many-shot slices} (denote as $M$): these have enough data and will not receive data augmentation. The example extrapolator is trained with $M$ only and used to infer new examples in $F$ despite never having seen any examples in $F$ during training.

It is important to note that we refer to ``few-shot'' to mean that there are slices of the data \emph{within} the task that have very few examples. The other notion of few-shot learning, where there are overall few examples for the entire task, is outside of the scope of our experiments.

\subsection{Example extrapolation (Ex2)}\label{sec:extrapolator}

\paragraph{Task definition.}

With a formal notion of slices, we can now define the example extrapolation task. First, let $p(e)$ denote the true underlying distribution over examples. And for a given slice $s$, let $p(e \mid s) \defeq p(e \mid \slices(e) = \mathtt{true})$ be the distribution of examples restricted to that slice.

In order to generalize to new, unseen slices, we featurize $s$ with a random sample of $K$ examples from slice $s$, denoted as $e_{1:K}$.
The example extrapolation task is to model the full distribution of slice $s$ given only those exemplars:
\begin{align*}
p(e \mid s) &= \pextrap(e \mid e_{1:K})
\end{align*}

When deciding how many exemplars $K$ to condition on, it is important to ensure that they are enough to illustrate the intra-slice variance; we expect that conditioning on a single exemplar will generally be insufficient. Section \ref{sec:analysis} explores varying the size of $K$.

\paragraph{Training procedure.}

Optimization of the example extrapolator is straightforward once we define the inputs and outputs.

Given a training set $D$, let $D_1, \ldots, D_s$ denote its $S$ different slices. Let $e_{1:K} \simwr D_s$ denote a sample of $K$ examples from $D_s$, drawn uniformly without replacement. Then the training objective is:
\begin{align*}
\sum_{s \in M} p(s) \sum_{e^{*} \in D_s} E_{e_{1:K} \simwr D_s \backslash e^*}[\log \pextrap(e^{*}\mid e_{1:K})]
\end{align*}
where the term $p(s)$ is a user-defined prior probability of each slice, which we estimate empirically from the training data in our experiments.

To optimize this objective, we iterate over all training slices ($s \in M$), and every example ($e^*$) in each slice. For each example, we sample $K$ other examples ($e_{1:K}$) from the same slice, excluding $e^*$ itself. We then optimize the log-likelihood of $e^*$ as output given $e_{1:K}$ as input.

\paragraph{Model implementation.}
We implement our example extrapolator as a neural sequence-to-sequence model. In particular, we use T5 \cite{Raffel2020ExploringTL}, a text-to-text Transformer model \cite{vaswani2017attention} that was pre-trained on a large text corpus. 
This provides the network with a large amount of world knowledge, which is crucial for the model's ability to \emph{extrapolate} beyond the given examples. For example, the last example in Table~\ref{tab:examples} requires extrapolating \nl{New Beaver} and \nl{Keeneland} from the input exemplars to \nl{Steven's Pass} in the output, which requires some world knowledge that pre-trained models are known to contain~\cite{petroni2019language,roberts2020much}. We show that this pre-training is crucial for an effective Ex2 model in Section~\ref{sec:analysis}.

\paragraph{Exemplar (de)serialization}
Since T5 operates over plain text inputs and outputs, we must represent the input exemplars $e_{1:K}$ and the output $e^*$ as text. For any given task, we assume the user provides a function $\totext$ that maps a single example to a string, and a function $\fromtext$ that maps a string back to an example.

An important subtlety in the $\totext$ function is whether the extrapolator is allowed to ``cheat'' when determining the boundaries of the slice. Suppose we are using Ex2 for text classification, with our data sliced by label, and suppose we specify the $\totext$ function to prepend the label name to the input sentence (e.g.\ ($x=$``play a song'', $y=$\texttt{PlayMusic}) is mapped to ``PlayMusic: play a song''). On the one hand, the model may be able to take advantage of the semantics of the label name, gleaned from pre-training. On the other hand, it will be easier for the extrapolator to determine the properties of the slice by memorizing the label and ignoring everything else.
This challenge is analogous to the task memorization associated with meta-learning algorithms~\cite{meta_memorization}, where leaking task-level information to the meta-learner results in poor generalization.

We hypothesize that the benefits of anonymization outweigh the losses, so we ensure that $\totext$ \emph{anonymizes} any slice information, and that $\fromtext$ can project the anonymized generation back to a fully realized example. Examples of the anonymization strategy for each task are shown in Table~\ref{tab:examples}. We explore this hypothesis empirically in Section \ref{sec:analysis}.

\subsection{Using Ex2 for data augmentation}\label{sec:augmentation}
Our example extrapolator enables us to take $K$ examples from a slice and generate additional examples from the same slice. Concretely, given a slice $D_s$, we sample $K$ exemplars without replacement, $e_{1:K} \simwr D_s$, feed them into the extrapolator, then randomly sample from the extrapolator:
\begin{gather*}
\textit{output-text} \sim \pextrap(\cdot \mid \totext(e_{1:K})) \\
\tilde{e} = \fromtext(\textit{output-text})
\end{gather*}
By repeatedly sampling in this fashion, we can produce an arbitrary number of new labeled examples, discarding any invalid ones that cannot be parsed by $\fromtext$.

Let $\tilde{D}_s$ denote all the new examples sampled from our extrapolator for under-represented or few-shot slice $s \in F$. We can then form a new, augmented training set, which we use to train the final downstream model:
\begin{align*}
\tilde{D} = D \cup \tilde{D}_F
\end{align*}
The amount of data generated for each slice is up to the user, but would ideally correct for the under-representation and reflect the true underlying distribution of the slices.

\paragraph{Analogy to distillation.}
For ease of discussion, we may also refer to the example extrapolator as the ``teacher'', and the downstream model as the ``student''. This terminology is deliberately reminiscent of model distillation~\cite{Tarvainen2017MeanTA}, where a ``teacher'' is used to label a large number of unlabeled inputs ($x$'s) to be consumed by a ``student''. The Ex2 approach is similar, except that the teacher does not label pre-existing $x$'s and instead synthesizes completely new $(x, y)$ pairs.

\section{Experiments}\label{sec:experiments}

To validate the generality of the Ex2 recipe, we evaluate our approach on a range of different language understanding tasks: text classification (a simple setup that resembles our running example), intent classification + slot-filling (a more complex task with a structured output space), and relation extraction (a highly multi-class problem with strong prior work in the few-shot setting). 

Across all three tasks, our results consistently show that a model trained with \textbf{Ex2} data augmentation outperforms our baselines. In the cases of SNIPS and especially relation extraction, where strong published baselines are available, we achieve a new state of the art. 

\paragraph{Data splits.} 
In our experiments, we explicitly designate certain slices of the dataset as few-shot and the others as many-shot. Furthermore, we define the \textbf{few-shot split} of a dataset $D_F$ to be the set of all examples belonging to a few-shot slice, and the \textbf{many-shot split} $D_M$ to be all other examples.
Table~\ref{tab:splits} gives the shorthand notation we use for these splits which are further sub-divided into Train, Development and Test.

\begin{table}[t!]
\newcolumntype{Y}{>{\centering\arraybackslash}X}
\newcommand{\colindent}{\;}
\setlength{\tabcolsep}{.35em}

\small
\centering
\begin{tabular}{l c c  c}
\toprule
                & Train & Dev. & Test \\
\midrule                
Many-shot split  & $D_{M, \text{train}}$ & $D_{M, \text{dev}}$ & $D_{M, \text{test}}$ \\
Few-shot split  & $D_{F, \text{train}}$ & $D_{F, \text{dev}}$ & $D_{F, \text{test}}$
\\
\bottomrule
\end{tabular}

\caption{Data splits used in Ex2 experiments.}
\label{tab:splits}
\end{table}

\begin{table*}[ht!]
\setlength{\tabcolsep}{0.3em}
\centering
\scriptsize
\def\tabularxcolumn#1{m{#1}}
\begin{tabularx}{\linewidth}{lXX}
\toprule
Task & Input sequence & Output sequence \\
\midrule
Relation Extraction 
&  \nl{An \cspan{head}{oil spill} caused by \cspan[light cyan]{tail}{a collision} closed the ship channel.} & \nl{\colorbox{light green}{effect}}\\
\cmidrule{1-3}
Classification
&   \nl{Are my tires under-inflated} &
    \nl{\colorbox{light green}{tire\_pressure}}\\
\cmidrule{1-3}
Slot filling 
&   \nl{How will the weather be at Steven's Pass this weekend} &
    \nl{\colorbox{light green}{GetWeather} | How will the weather be at \cspan{location}{Steven's Pass} \cspan[light cyan]{time}{this weekend}}\\
\bottomrule
\end{tabularx}
\caption{Training examples for the T5 student models. Span names and intents are highlighted.}
\label{tab:student_ex}
\end{table*}

For relation extraction, prior work had already designated certain slices as few-shot --- we consider the same ones for direct comparison. For intent classification/slot-filling, we \textbf{cross-validate} by running one experiment for each slice in the dataset, where that slice is designated the few-shot one and its training set is artificially truncated to $K$ examples. In all cases, the Train/Dev/Test axis of our splitting follows the original benchmarks.

\paragraph{Evaluation.}
When reporting downstream student model performance, we consider both \textbf{Overall} performance (averaging across $D_{M} \cup D_{F}$) and \textbf{Few-shot} performance (averaging only over $D_{F}$). Tables in this section report the overall and few-shot test performance.

\paragraph{Baselines.}
The output of Ex2 is simply additional synthetic data, which must then be consumed by the downstream student model. To measure the contribution of this additional data, we always compare between the same student configuration.\footnote{
We use the \emph{overall} accuracy of $D_{M, \text{dev}} \cup D_{F, \text{dev}}$ for early stopping for FewRel, and overall macro F1 for the other tasks.} The only difference between the following setups is the data that the student is trained on:

\begin{enumerate}[noitemsep,topsep=0pt,parsep=0pt,partopsep=0pt]
    \item \textbf{Baseline}: The student only trains on the original data without any augmentation ($D_{M, \text{train}}~\cup~D_{F, \text{train}}$).
    \item \textbf{Upsampled}: The student trains on original data ($D_{M,\text{train}}~\cup~D_{F,\text{train}}$), but the examples from the few-shot slices $D_{F,\text{train}}$ are up-sampled to match the median frequency of the many-shot slices.
    \item \textbf{Ex2}: The teacher is trained on the many-shot training data ($D_{M, \text{train}}$).\footnote{We use the token accuracy on $D_{M,\text{dev}}$ for early stopping.} Synthetic data for the few-shot slices $\tilde{D}_{F}$  are sampled to match the median frequency of the many-shots slices. The student trains on the union of original data and synthetic data ($D_{M,\text{train}}~\cup~D_{F,\text{train}}~\cup~\tilde{D}_{F}$).
\end{enumerate}
All other aspects of the model are held fixed across these setups. 
When previously published results for a task are available, we also compare against other model types.

\paragraph{Model architectures.}
For simplicity, we use T5~\cite{Raffel2020ExploringTL} as our student models here, since they achieve state-of-the-art performance even without any data augmentation. Table \ref{tab:student_ex} shows how each task is cast in the seq2seq framework.
We present results where both the teacher and student models are finetuned from T5-XL\footnote{We use the T5.1.1 version that is only pretrained on unlabeled data~\cite{roberts2020much}. The teacher models are finetuned for 3 epochs for FewRel and 10 epochs for CLINC150/SNIPS. The student models are finetuned for 10$k$ steps for FewRel and 20$k$ for the others. All models use batch size of 128. All other hyper-parameters are set to T5's default.} unless otherwise noted. We also evaluate the impact of T5 model sizes in Section~\ref{sec:analysis}. 

\begin{table}[t!]
\newcolumntype{Y}{>{\centering\arraybackslash}X}
\newcommand{\colindent}{\;}
\setlength{\tabcolsep}{.4em}

\footnotesize
\centering
\begin{tabular}{l c c c c}
\toprule
& \multicolumn{2}{c}{Overall} & \multicolumn{2}{c}{Few-shot} \\
    \cmidrule(lr){2-3}\cmidrule(lr){4-5}
  &  Acc. & Macro F1 & Acc. & Macro F1\\
\midrule
Baseline  & \textbf{97.4} & 95.3 & 93.7 & 60.6   \\
Upsampled  & \textbf{97.4} & 95.0 & 94.4 & 64.5   \\
Ex2  & \textbf{97.4} & \textbf{96.1} & \textbf{95.6} & \textbf{80.4}   \\
\bottomrule
\end{tabular}

\caption{Accuracy of CLINC150 classification task on the official test set averaged across 10 held-out domains.}

\label{tab:clinc150_main}
\end{table}

\subsection{Text Classification}

Our first task illustrates one of the simplest applications of Ex2. Given a short text snippet such as ``play a song'', a text classifier must select the correct label (e.g., \texttt{PlayMusic}). For this task, we evaluate on the CLINC150 dataset~\cite{Larson2019AnED}.
The original dataset contains 10 domains with 15 class labels per domain and 100 training examples per class label (a total of 15,000 examples).\footnote{We did not use the out-of-scope portion of the dataset.}
We use the cross-validation setup and report results averaged over 10 runs, where each run chooses a different domain to contain few-shot slices.

For Ex2, we slice the dataset by class label, and set the number of exemplars to be $K=10$. For the T5 student model, the input text to T5 is simply the plain text snippet, and the output is the string representation of the label (See~Table \ref{tab:examples} for Ex2 input-output pairs).

\paragraph{Results.}
Table~\ref{tab:clinc150_main} shows the accuracy and macro F1 results on both the overall and the few-shot splits. Ex2 significantly improves over the upsampled baseline on the few-shot slices (+15.9 ppt in terms of macro F1), while maintaining the same performance on the overall accuracy.\footnote{Some previous works on few-shot intent classification of CLINC150~\cite{Zhang2020DiscriminativeNN} use the setup where all intents are few-shot, therefore our results are not directly comparable.}

\subsection{Intent Classification and Slot Filling}

Intent classification is the task of mapping a user utterance to an intent label, as above. Slot filling is the task of identifying argument spans of the intent within the utterance.
We use the SNIPS~\cite{coucke-etal-2018-snips} dataset,\footnote{We use the preprocessed version from~\citet{goo2018slot} at https://github.com/MiuLab/SlotGated-SLU.} which contains 7 intents (domains) with a total of 39 different slot types. 

For Ex2, we slice the data by intent label and set the number of exemplars to be $K=10$. When truncating $D_{F, \text{train}}$, we use a greedy algorithm\footnote{The algorithm is inspired by \citet{yang-katiyar-2020-simple} to ensure that all slot types are present in the smaller set. First, we identify the slot type present in the slice but least well-attested in the current set $F_{\text{train}}$ (with ties broken in favor of the more infrequent type). We then randomly select an exemplar containing that slot type from the domain. For this purpose, exemplars with no slots are assumed to have a single \texttt{null} slot. This ensures that the teacher and student both have access to a maximally complete and diverse set of inputs.} to select source exemplars such that each one is guaranteed to share a slot type with the target. 

For the T5 student model, the input to T5 is the plain text utterance, and the output is the same plain text utterance, except prefixed with the predicted intent, and with special tokens inserted to mark the beginning and end of slot values (cf.\ Table~\ref{tab:student_ex}).

\begin{table}[t!]
\newcolumntype{Y}{>{\centering\arraybackslash}X}
\newcommand{\colindent}{\;}
\setlength{\tabcolsep}{.25em}

\footnotesize
\centering
\begin{tabular}{l c c c c}
\toprule
 & \multicolumn{2}{c}{Overall} & \multicolumn{2}{c}{Few-shot} \\
         \cmidrule(lr){2-3}\cmidrule(lr){4-5}
  & Intent & Slot  & Intent & Slot\\
\midrule
\citet{kumar2019closer}$^*$ & 95.9 & -- & -- & --  \\
\citet{Krone2020LearningTC}$^*$ & -- & -- & 88.9 & 62.1  \\
\citet{Hou2020FewshotST}$^*$ & -- & -- & -- & 75.0  \\
\cmidrule(){1-5}
Baseline & 95.2 & 93.0 & 74.0 & 70.0  \\
Upsampled &  95.9 & 92.7 & 80.0 & 69.5  \\
Ex2 & \bf 97.8 & \bf 93.5 & \bf 94.0 & \bf 75.3   \\
\bottomrule
\end{tabular}

\caption{Intent accuracy (Intent) and micro slot F1 (Slot) on the SNIPS dataset. The numbers are from the official test set and averaged across all the 7 domains. 
$^*$: Prior results are not strictly comparable due to difference in data sampling strategies and training setup.
}
\label{tab:snips_main}
\end{table}

\paragraph{Prior results.}
\citet{kumar2019closer} evaluate a data augmentation technique for few-shot intent classification on the SNIPS and TOP datasets. Their approach involves permuting sentence embeddings $D_{F, \text{train}}$ set (across a variety of different permutation functions), and training the system on the permuted embeddings in addition to the original embeddings. The approach is restricted to sentence classification, however.

\citet{Hou2020FewshotST} and \citet{Krone2020LearningTC} both involve explicitly aligning token- or span-vectors from an incoming query to prototype vectors derived from $F_{\text{train}}$ and computing the similarity between them directly. 

\citet{kumar2019closer} and \citet{Hou2020FewshotST} use BERT \citep{devlin2019bert} to encode queries, whereas \citet{Krone2020LearningTC} found ELMo \citep{peters2018deep} to work better for this task in their experiments.

\paragraph{Results.}

Table \ref{tab:snips_main} shows how our system compares to the simple T5 baseline with and without upsampling. It can be observed that upsampling the few-shot classes improves intent accuracy over the baseline, but its impact on slot-filling is considerably more modest.
Ex2, however, drastically improves intent accuracy while also increasing slot F1 (by 20 ppt. and 5 ppt. respectively) on the few-shot slices.
These improvements in the few-shot domain appear to carry over into the overall scores, as evidenced by a 2.5 ppt.\ increase in overall intent accuracy and a 0.5 ppt.\ increase in overall slot F1.

We also include previous published results on SNIPS, but they only serve as a rough reference to demonstrate that T5 is a competitive baseline, since there are slight differences in the experimental setup. The numbers from~\citet{kumar2019closer},  \citet{Hou2020FewshotST} and \citet{Krone2020LearningTC} are not strictly comparable to ours, because they use a different data truncation strategy, and a different train/development setup\footnote{\citet{Hou2020FewshotST} truncate the few-shot domain to have close to $5$ instances of each \emph{slot} type rather than $10$ instances of each \emph{intent} type. They also use one domain for development in cross-validation, whereas \citet{kumar2019closer} did not include $D_{F,dev}$ their development set.}. 

Despite the strong empirical results over baselines, we find that the quality of the synthetic examples is noticeably worse than in the other tasks, with the training intents sometimes ``bleeding'' into the few-shot intent (e.g.\ $\tilde{e}=$ (``play me something close by neylandville'', \texttt{BookRestaurant}), with bleeding from the \texttt{PlayMusic} intent). In the SNIPS dataset, there are only 7 slices of data from which the Ex2 teacher can learn (an order of magnitude fewer than the other datasets); we infer from this that it is important to have a large number of slices so that Ex2 can reason by analogy rather than memorize the many-shot slices.

\subsection{Relation Extraction}
In relation extraction, a model is given a passage of text featuring two entity mentions, and must predict the relation between the pair of entities.

We evaluate on the well-studied few-shot relation extraction benchmark, FewRel dataset~\cite{han2018fewrel}, where some relations are designated for few-shot learning. Previous results have reported super-human performance on FewRel~\cite{soares2019matching}. However, the original task only requires the model to select the correct relation from a pruned set of possible options, rather than the full catalogue of relations.

We therefore use a more challenging variant of FewRel (FewRel-Open), where the model must choose from all relations (and in the case of nearest neighbor models choose from all training neighbors). This setup is much closer to real-world applications of relation extraction and explicitly evaluates the models ability to predict under-represented relations while being overwhelmed by a highly-unbalanced prior in the training data.

The 64 Wikipedia training relations with 70k sentences are used for teacher and student training. In addition to in-domain Wikipedia evaluation, we also evaluate on out-of-domain generalization with the NYT, SemEval, and PubMed evaluation sets from FewRel 2.0~\cite{fewrel2} and report the macro average over all domains.

For Ex2, we slice the dataset by relation label, and treat the few-shot relations defined in the original FewRel dataset as our underrepresented slices. We set the number of exemplars to be $K=5$. For the student model, the input text and entity mentions are formatted into a plain text by marking the start and end of each entity mention using special tokens. The text output from T5 is the string name of the relation (see Table~\ref{tab:student_ex}).

\paragraph{Prior Results.}
In addition to the data augmentation baselines described earlier, we compare to the state-of-the-art Matching the Blanks (MTB) model~\cite{soares2019matching}, which is a nearest-neighbor approach based on BERT. MTB was trained with an unsupervised objective that aims to improve the modeling of entity relations. 

\begin{table}[t!]
\newcolumntype{Y}{>{\centering\arraybackslash}X}
\newcommand{\colindent}{\;}
\setlength{\tabcolsep}{.4em}

\footnotesize
\centering
\begin{tabular}{l c c }
\toprule
& Overall Acc. & Few-shot Acc. \\
\midrule
 MTB 	    & 68.6  &	50.4. \\
 \cmidrule(lr){1-3}
Baseline & 77.3  &	64.5 \\
Upsampled   & 75.8	&	62.5   \\
Ex2         &  \textbf{78.0}	&	\textbf{70.7} \\
\bottomrule
\end{tabular}
\caption{FewRel-Open results on the test split, averaged over the Wiki, NYT, SemEval, and PubMed domains.}
\label{tab:fewrel_main}
\end{table}

\paragraph{Results.}
The first notable result is that while MTB exceeds human performance on the original FewRel task, the accuracy of MTB drops dramatically in the more challenging and realistic FewRel-Open task. It achieves an average few-shot accuracy of 69\% in the overall evaluation and 50.5\% when evaluating only on examples with the few-shot labels. We hypothesize that teasing apart gold and random distractor neighbors is easy, but avoiding distractors from an entire training set worth of potential neighbors is much more challenging.

Interestingly, we found that our no-data-augmentation T5 baseline already improves over MTB, even though it does not employ a custom architecture specifically designed to improve few-shot learning. This could simply be attributed to the larger size of T5-XL compared to MTB, which is based on BERT-large. Since we aim to compare to the best-performing baseline, we mainly compare to the T5 baseline.

When we perform data augmentation with Ex2, we observe another significant improvement in accuracy, setting a new state of the art for both few-shot relations (7.2 ppt increase) and the overall accuracy (2.2 ppt increase).
\section{Analysis}\label{sec:analysis}

\subsection{Ablations}
Ex2 relies on three intuitions that we aim to justify empirically in this section: 
\begin{enumerate}[noitemsep,topsep=0pt,parsep=0pt,partopsep=0pt]
\item It is critical to have a broad range of source exemplars in order to show the model the boundaries of the data slice under consideration.
\item The identity of the slice should be obfuscated in order to encourage the model to infer the slice distribution using the source exemplars.
\item The model needs access to world knowledge that is not present in the training data in order to generate accurate and diverse outputs.
\end{enumerate}
We present ablations that test these three claims. The experimental setups for these analyses are identical to those presented in the main experiments, except we present results on the validation sets.

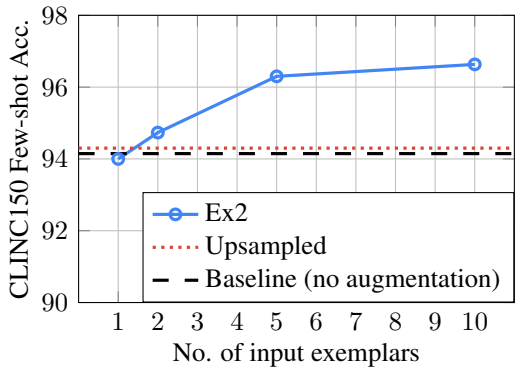
\begin{figure}[t!]
\begin{tikzpicture}
\begin{axis}[
   width=0.95\columnwidth,
   height=0.7\columnwidth,
   legend cell align=left,
   legend style={at={(1, 0)},anchor=south east,font=\small},
   mark options={mark size=2},
   font=\small,
   xmin=0, xmax=11,
   ymin=90, ymax=98,
   ytick={90, 92, 94, 96, 98},
   xtick={1, 2, 3, 4, 5, 6, 7, 8, 9, 10},
   ymajorgrids=true,
   xmajorgrids=true,
   xlabel style={yshift=0.5ex,},
   xlabel=No. of input exemplars,
   ylabel style={align=center},
   ylabel=CLINC150 Few-shot Acc.,
   ylabel style={yshift=-0.5ex,}]
    \addplot[mark=o, g-blue, line width=1.2pt] plot coordinates {
      (1, 94.00000095367431)
      (2, 94.73333299160004)
      (5, 96.30000054836273)
      (10, 96.63333277702332)
    };
    \addlegendentry{Ex2}
    \addplot[g-red, line width=1.2pt, dash pattern=on \pgflinewidth off 2pt] plot coordinates {
      (0, 94.30033358764648)
      (11, 94.30033358764648)
    };
    \addlegendentry{Upsampled}
    \addplot[black, line width=1.2pt, dash pattern=on 6pt off 6pt] plot coordinates {
      (0, 94.14814843071831)
      (11, 94.14814843071831)
    };
    \addlegendentry{Baseline (no augmentation)}
\end{axis}
\end{tikzpicture}
\vspace{-1em}
\caption{Ablating the number of source exemplars. For text classification (CLINC150), Ex2 with only one input exemplar reduces to paraphrasing data augmentation, which does not improve over the baselines.}
\label{fig:mask}
\end{figure}
\paragraph{Size of $K$}
We use CLINC150 to demonstrate the importance of jointly reasoning across different exemplars by varying the number of exemplars $K$. We choose this intent classification task because the special case where $K=1$ reduces to a paraphrasing data-augmentation approach. Since a paraphraser only observes one exemplar, it cannot reason about the different axes of variance in a slice, and only has enough information to generate a generically \emph{similar} example.

As expected, Figure~\ref{fig:mask} shows that the paraphrasing special case does no better than the baselines. Using just $K=2$ exemplars already improves the few-shot accuracy above the baseline, and we observe substantial improvement with even more exemplars. Note that in all of these settings, the teacher performs inference on the same amount of few-shot data, and $K$ only controls the number of exemplars that the teacher encodes at the same time. Therefore, these results demonstrate the importance of \emph{cross-exemplar} reasoning in Ex2.

\paragraph{Anonymization strategy}
In this experiment, we compare our original Ex2 model with ones that lack slice anonymization; we use the SNIPS dataset for this experiment because it includes both classification and slot-filling subtasks, meaning there are two ways to anonymize the data. Table \ref{tab:anon} compares Ex2 and baselines to two non-anonymized models: one that includes slot label names and another that also prepends the intent name to the source sequence.

The hypothesis appears to be borne out to some extent: the anonymized Ex2 models outperform the non-anonymized ones in terms of few-shot intent accuracy. Surprisingly, argument F1 is lower than in the non-anonymized models,\footnote{This pattern held even after a second trial of this experiment. In Ex2-L models, anonymization improves intent accuracy dramatically and is uncorrelated with argument F1.} indicating that providing slot and/or intent names improves argument synthesis. It's likely that label strings (such as \texttt{artist} or \texttt{AddToPlaylist}) provide some semantic signal that extra-large networks can take advantage of, and that it's easier to connect the semantics of the label to the semantics of possible fillers than to whole queries. This points to a tradeoff between providing the model with information it can use to generalize and withholding information that it may memorize.

\begin{table}[t!]
\newcolumntype{Y}{>{\centering\arraybackslash}X}
\newcommand{\colindent}{\;}
\setlength{\tabcolsep}{.25em}
\footnotesize
\centering
\begin{tabular}{ l c c c c}
\toprule
& \multicolumn{2}{c}{Overall} & \multicolumn{2}{c}{Few-shot} \\
 & Intent & Slot  & Intent & Slot \\
\midrule
Baseline & 96.0 & 92.9 & 78.4 & 72.2  \\
Upsampled & 96.6 & 92.5 & 82.1 & 70.7  \\
\cmidrule(){1-5} 
Ex2 (anonymized) & \bf 98.5 & 93.2 & \bf 96.6 & 76.7  \\
Ex2 (slot names) & \bf 98.5 & 93.3 & 95.7 & 78.0  \\
Ex2 (slot \&{} intent names) & \bf 98.5 & \bf 93.6 & 95.9 & \bf 79.4  \\
\bottomrule
\end{tabular}
\caption{Intent accuracy (Intent) and argument micro-F1 (Slot) on the SNIPS dataset, comparing Ex2-XL teachers that use full anonymization, include slot labels, and include both slot and intent labels. Anonymization improves intent classification but hurts slot F1.}
\label{tab:anon}
\end{table}

\begin{table}[t!]
\newcolumntype{Y}{>{\centering\arraybackslash}X}
\newcommand{\colindent}{\;}
\setlength{\tabcolsep}{.5em}

\footnotesize
\centering
\begin{tabular}{lcc}
\toprule
& \makecell[c]{Overall\\Accuracy} & \makecell[c]{Few-shot\\Accuracy} \\
\midrule
None (Baseline) & 77.9 & 65.7\\
Upsampled & 76.2 & 62.5\\
 \cmidrule(){1-3}
Ex2-XL & 80.4 & 69.2 \\
Ex2-L & 76.6  & 63.5\\
Ex2-Base & 72.6 & 55.3 \\
Ex2-XL random init. &  68.2 & 46.2\\
\bottomrule
\end{tabular}
\caption{Impact of Ex2 model size and pre-training. ``random init'' refers to initializing the parameters of the Ex2 teacher randomly, without T5 pre-training. For all rows, the student model is T5-XL. Pre-trained capacity is positively correlated with accuracy.}
\label{tab:pretraining}
\end{table}

\paragraph{Pre-training}
We train an Ex2 model from scratch and compare it to one that has been fine-tuned from a T5 model. We evaluate this on FewRel, which requires synthesizing the longest and most complex examples out of the three tasks in this paper. Results in Table \ref{tab:pretraining} demonstrate that a randomly initialized Ex2 is completely ineffective, with the generated examples introducing substantial noise into the system with little tangible gains. Furthermore, we observe a correlation between model size and performance; a sufficiently large pre-trained model (at least T5-XL) is necessary for Ex2 to be effective for FewRel. As stipulated in Section~\ref{sec:approach}, this suggests the world knowledge from pre-training is critical to the ability of Ex2 to \emph{extrapolate} to new examples containing of new concepts rather than simply recombining or paraphrasing existing parts from the input exemplars. 

\newcommand{\bigcell}[1]{\begin{tabular}[t]{@{}l@{}}#1\end{tabular}}
\begin{table*}[t!]
\setlength{\tabcolsep}{0.8em}
\centering
\scriptsize
\def\tabularxcolumn#1{p{#1}}
\begin{tabularx}{\linewidth}{XX}
\toprule
Exemplars & Sampled Generations  \\
\midrule
(1) can you help me improve my spanish | i want to learn french | please help me with my english grammar | learn new chinese words | i would like to learn some japanese
    & \bigcell{
    best way to learn spanish
    \\ can i improve my japanese language
    \\ i need help with my italian language
    }\\
\cmidrule(){1-2}    
(2) can you help me improve my spanish | i want to learn french | please help me with math | i want to learn ap physics | teach me to write java please  
    & \bigcell{
    can you teach me history
    \\ please help me learn c
    \\ please teach me to count
    }\\
\cmidrule(){1-2}
(3) show me the latest wug |
can you bring up another wug |
open the wug list |
what are the new \emph{updates} |
delete my wug 
    & \bigcell{
    please show me the wug list
    \\ please tell me what \emph{wugs} are there
    \\ what is the latest wug
    }\\
\cmidrule(){1-2}
(4) start the wug machine |
wug my living room please |
wug all the rooms tomorrow morning |
stop \emph{cleaning} |
wug it again 
    & \bigcell{
    get cleaning
    \\ let's start \emph{wugging}
    \\ i gotta clean my bedroom
    }\\
\bottomrule
\end{tabularx}
\caption{Selected batches of sampled generations from an Ex2 teacher trained on CLINC150. (1), (2): we can control whether Ex2 generates new languages or new subjects by controlling the variations in the input exemplars. (3), (4): the model generalizes to the plural or new tenses of ``wug'' by composing with other exemplars in the input (``updates'' and ``cleaning'').}
\label{tab:cluster_width}
\end{table*}

\subsection{Qualitative analysis of Ex2 outputs}
We posit that Ex2 is able to effectively use the source exemplars to estimate the boundaries of the intended slice when synthesizing a new example. In Table \ref{tab:cluster_width} we demonstrate this qualitatively. The first column shows sets of five exemplars passed to an Ex2 model trained on CLINC150 (with ``auto'' as the held-out domain), and the second shows three different outputs synthesized from each set\footnote{We generate synthetic outputs by batches of 3, and show the selected batches here.}.

When comparing examples (1) and (2) --- which differ only in the specificity of the slice, with (1) representing queries about help learning languages and (2) representing queries about help learning academic subjects more broadly --- the generated examples stay confined to the regions specified by the source exemplars while not repeating any of the source queries.

Examples (3) and (4) show that not only can Ex2 learn the boundaries of clusters, it can pass a variation of the ``wug test'', using context to infer the semantic and morpho-syntactic category of nonce words with previously unseen meanings. We see that Ex2 can compose new syntactic forms based on variations in the exemplars. When observing a word such as \emph{updates} or \emph{cleaning} that fills the same semantic role as \emph{wug} in other source exemplars but with different morphology, Ex2 is more likely to generate an example using the word \emph{wug} that bears the same form. This demonstrates an extreme case of out-of-domain generalization, where Ex2 can be used to quickly adapt to new or even conflicting information.
\section{Related Work}\label{sec:related}

\subsection{Data augmentation}

There is a large body of research on data augmentation~\cite[inter alia]{jia2016data,Andreas2020GoodEnoughCD,Akyrek2020LearningTR}.
Within this literature, our approach is most related to recent work on data augmentation for NLP using pre-trained language models (LMs):~\citet{kumar2019closer, anaby2020not} perform data augmentation for text classification by fine-tuning an LM to synthesize new inputs $x$ for a given label $y$ --- modeling $p(x | y)$.
Like these approaches, Ex2 uses LM pre-training to acquire world knowledge, and then fine-tunes the LM to perform data generation. But our generation task is notably different: prior work conditioned the data generator on an output label $y$, whereas Ex2 conditions on a collection of exemplars $[(x_1, y_1), \ldots, (x_K, y_K)]$.

This yields several advantages. First, it enables us to generate examples for new slices that were never seen at training time, since the extrapolator can reason \emph{by analogy} instead of memorizing the identity of labels. Second, it allows us to perform data augmentation along dimensions other than the output label --- exemplars can be used to express any desired quality (e.g., a particular sentence length or syntactic structure), not just a desired label. This makes Ex2 applicable to tasks beyond classification. Finally, note that Ex2 synthesizes entirely new labeled examples ($(x, y)$ pairs), rather than just the $x$. This allows Ex2 to naturally cover variation in the output space, which is essential for tasks with large and compositional output spaces such as parsing.

\subsection{Few-shot learning with language models}

Beyond data augmentation, large language models have been used in various other ways to address few-shot learning~\cite{Schick2020ExploitingCQ,Brown2020LanguageMA}. Our approach is most related to the in-context learning approach of GPT-3~\cite{Brown2020LanguageMA}. Similar to Ex2, GPT-3 also conditions on a collection of exemplars.

However, the two models solve different tasks. GPT-3 maps an input $x$ to an output $y$, whereas Ex2 generates a new $(x, y)$ pair. In other words, Ex2 uses a large LM to generate data, whereas GPT-3 uses a large LM as the model itself. Using large LMs for data generation rather than direct inference has practical benefits: data can be inspected and cleaned by humans, easily persisted, and finally used to train much smaller models that are cheaper to deploy than a large LM.\footnote{A model like GPT-3 could also be used for data generation, by using it to label a large number of unlabeled $x$'s --- as done in distillation. But in many NLP tasks (e.g., natural language inference), coming up with a valid $x$ is non-trivial, and often even harder than predicting the label.}

The purpose of exemplars is also different: for GPT-3, exemplars are used to describe the overall task (and hence drawn uniformly from the training set), while for Ex2, exemplars are used to describe a particular slice of the task. This distinction is important for tasks with many slices. For example, consider a few-shot document classification problem with 1000 possible labels (where each label is a slice), and we have 5 examples for each label. Using Ex2, we would condition on $K=5$ exemplars at a time to generate new examples. In contrast, GPT-3 requires one set of exemplars to describe the entire task, so it must condition on at least $K=1000$ exemplars to ensure that every label is included at least once in the set. This becomes computationally intractable.

On the other hand, it is attractive that GPT-3 generalizes over many tasks, whereas Ex2 only targets a single task. In future work, one could imagine using Ex2 to generalize across tasks by grouping multiple tasks together, and learning over the union of all their slices.

Lastly, Ex2 is fine-tuned to perform few-shot data augmentation, whereas GPT-3 is not fine-tuned. Therefore, GPT-3 users must be careful to format examples in a way that resembles ``natural'' text encountered during pre-training -- such ``format engineering'' can greatly affect performance~\cite{shin2020eliciting,Schick2020ExploitingCQ}. In contrast, fine-tuning allows Ex2 to introduce arbitrary formats and annotations that deviate from natural language, which is necessary for slice anonymization and modeling more structured tasks.

\subsection{Nearest neighbor methods}

Among methods for few-shot learning, nearest-neighbor and other instance-based models constitute another prominent category that conditions on a collection of examples~\cite{vinyals-etal-2016-matching,snell-etal-2017-prototypical,Sun2019HierarchicalAP,yang-katiyar-2020-simple,Hou2020FewshotST,ziyadi-etal-2020-example}.

It is worth noting that instance-based models require modest specialization, since inputs must be encoded into feature vectors, whereas Ex2 is model-agnostic. In fact, they are mutually compatible approaches that aim to improve few-shot learning in complementary ways.
\section{Discussion}
We address several potential concerns about the use of synthetic data generated from a highly expressive neural model.

\paragraph{Hallucination}
Ex2 is likely to generate text that is factually incorrect. While this initially sounds undesirable, we argue that for most tasks, the role of the downstream model is to understand language, not evaluate world knowledge. Therefore, an ideal model should be constrained to behave well on these hallucinated data points. For example, consider using Ex2 for a new relation indicating that entity 0 is the direction in which entity 1 sets. A robust relation extractor should predict that this relation exists in all of the examples below, regardless of world knowledge:
{
\small
\begin{itemize}\setlength\itemsep{0ex}
\item \nl{The [1 sun] sets in the [0 west]}
\item \nl{The [1 sun] sets in the [0 east]}
\item \nl{The [1 sun] sets in the [0 north]}
\item \nl{The [1 sun] sets in the [0 south]}
\end{itemize}
}

Ensuring that models make decisions via language understanding rather than memorizing facts or entities has been argued for named entity recognition~\cite{ner_eval} and coreference resolution~\cite{nec}.

\paragraph{Transparency}
Ex2 can also be considered a method for increasing the transparency of using large pre-trained LMs. The typical use of pre-trained LMs involves simply fine-tuning on the data and hoping that the model generalizes to new inputs. With Ex2, however, we would explicitly generate data that better cover the input space. While the new examples may contain mistakes (in the same way that a purely discriminative model would make mistakes), it would more transparently expose the regions where they happen.

\paragraph{Human curation}
While we argue that hallucination is not necessarily a problem, there are certainly cases where it is undesirable. Ex2 \emph{should not be used} in production-level models without making the most of Ex2's transparency by vetting the generated examples with human supervision. The most effective combination uses Ex2 to thoroughly cover possible variations (that may be tedious or difficult for humans) and uses human supervision to curate high-precision data.
\section{Conclusion}
We propose an approach for data augmentation by learning a neural example extrapolator (Ex2) that generates new labeled examples from a small sets of existing examples coming from the same ``slice'' of the dataset. 
Ex2 learns from slices of data with many data points, and uses that knowledge to synthesize new examples for slices of the data with few data points. We show that this is an effective approach for few-shot text classification, intent classification + slot filling, and relation extraction.

For future work, we hope to expand this approach to broader notions of slices, including slicing by languages for multilingual applications, slicing by tasks, or working with tasks that contain orders of magnitude more slices (e.g. entity linking). We also plan to explore whether Ex2 can be generalized to other modalities, such as images or speech, where we would need to explore architectures other than pre-trained seq2seq models. Finally, we believe that investigating the best way in which human supervision should be injected into applications of Ex2 is an important direction.

\section{Acknowledgements}
We thank Ice Pasupat, Yuan Zhang, Emily Pitler, Kristina Toutanova, Arun Chaganty, Zhuyun Dai, Terry Koo, Sebastian Ruder, Siamak Shakeri, Iulia Turc, and the Google Research Language team for their helpful feedback and discussions.

\bibliography{tacl2018}
\bibliographystyle{acl_natbib}

\end{document}